\pdfoutput=1
\documentclass[3p,12pt,nopreprintline]{elsarticle}
\usepackage{amssymb}
\usepackage{circledsteps}
\usepackage{float} 
\usepackage{color}
\usepackage{amsfonts}
\usepackage{amsmath}
\usepackage{pdflscape}
\usepackage{graphicx}
\usepackage{pdfpages}
\usepackage{subcaption}
\usepackage{bbm,bm}
\usepackage{stmaryrd}
\usepackage{booktabs}
\usepackage{yhmath}
\usepackage{flexisym}
\usepackage{adjustbox}
\usepackage{physics}
\usepackage{boldline,multirow}
\graphicspath{{./figures/}}
\usepackage{tabularx}
\usepackage{upgreek}
\usepackage[linesnumbered,ruled,vlined]{algorithm2e}
\usepackage[colorlinks,bookmarksopen,bookmarksnumbered,citecolor=blue,urlcolor=red]{hyperref}
\date{}

\begin{document}

\begin{frontmatter}

\title{Quantifying roof damage after wind hazards: two-phase transformer-based segmentation of post-event aerial imagery}

\author[1]{Weihe Sun\fnref{eq1}}
\author[1]{Chuan Yu\fnref{eq1}}
\author[2]{Kevin Zhao}
\author[3]{Eshan Shrestha}
\author[1,4]{Guangzhao Chen}
\author[4]{Douglas A. Smith}
\author[1,4]{Yimin Lu\texorpdfstring{\corref{cor}}{}}
\ead{Yimin.Lu@ttu.edu}
\cortext[cor]{Corresponding author}
\address[1]{Department of Civil, Environmental, and Construction Engineering, Texas Tech University, USA}
\address[2]{Lubbock High School, Lubbock Independent School District, USA}
\address[3]{Department of Computer Science, Texas Tech University, USA}
\address[4]{National Wind Institute, Texas Tech University, USA}
\fntext[eq1]{These authors contributed equally to this work.}

\begin{abstract}

Regional-scale quantification of roof damage after wind hazards is time-sensitive and critical for decision making, but indistinct roof boundaries and occlusion often complicate damage measurement from post-event imagery. Conventional building-by-building surveys take weeks to months, while state-of-the-art automated image analysis mostly assigns one damage label per building, which identifies damaged buildings without measuring the extent of roof damage. This study presents a two-phase semantic segmentation framework that measures roof damage from post-event aerial imagery alone: one Mask2Former model delineates the roof footprint, and the other one labels every roof pixel as intact, damaged, tarp-covered, or shadowed, yielding the damaged roof area and the damage ratio (damaged portion within the roof) per image tile. For the 2011 Joplin tornado dataset, the predicted damage ratio deviated from manual annotation by 6.66\% on average, and 43 of 47 image tiles with fully damaged roofs were correctly identified despite visually indistinct roof--debris boundaries. This deviation was 1.5 to 2 times as large for 3 alternatives: a single-phase model lost destroyed roofs that resemble debris, CLIPSeg struggled with tree shadows, and U-Net++ broke roofs into fragments. Fine-tuning with fewer than 100 labeled image tiles adapted the framework to Hurricane Melissa (2025) imagery from Jamaica with different building styles and reduced the error in damaged roof area by 87\%. This work advances image-based roof assessment from building-level damage detection to pixel-level damage quantification, which supports rapid regional-scale recovery decisions and provides a basis transferable to other hazards and structures.

\end{abstract}


\end{frontmatter}

\section{Introduction}

Roof damage is a primary form of building damage in wind hazards such as tornadoes and hurricanes \cite{Minor2005BuildingEnvelopeWindstorms}. How much of a roof is damaged matters as much as whether it is damaged. After an event, decision makers need the damaged roof area and the roof damage ratio across the entire affected region to rank repair needs and to allocate temporary roofing and recovery resources \cite{Kucharczyk2025RoofDamage}. However, these quantities are still obtained mainly through building-by-building field surveys, and one team may document about 20 to 100 buildings per day \cite{Womble2008WindDamage}. At this rate, the 7,964 buildings damaged by the 2011 Joplin tornado \cite{Kuligowski2014JoplinTornado} would require roughly 80 to 400 team-days. As a result, regional coverage takes weeks to months, exposes survey personnel to hazardous sites \cite{Gupta2019xBD}, and is rarely complete before cleanup and temporary repairs alter the damage scene \cite{Womble2008WindDamage}. In contrast, aerial imagery is usually acquired over the whole affected region within days of an event \cite{Joyce2009SatelliteHazards,roueche2022hybrid}, but its interpretation remains manual: virtual assessments have been reported to take 10 to 15 minutes per building \cite{roueche2022hybrid}, and a single incident can produce tens of thousands of images \cite{Scheele2024LADIv2}. The obstacle to rapid, regional-scale quantification of roof damage is no longer the acquisition of imagery but its automated conversion into damaged roof area and roof damage ratio.

Most computer vision methods for post-hazard imagery focus on the building level and assign one label to each building. Binary classifiers separate damaged from undamaged buildings \cite{Bouchard2022TransferLearning,Cao2018BuildingDamage,Cao2020Geolocation,Li2022BuildingDamage}, and multi-class models assign each building a damage grade or damage type, from pre- and post-event image pairs \cite{Gupta2019xBD,Shen2022BDANet,Kaur2023DAHiTrA} or from post-event imagery alone \cite{Miura2020BlueTarp,AlShafian2025DamageScope,xiao2026damage}. Even when the output is a segmentation map, every pixel of a building carries the label of that building \cite{Shen2022BDANet,Kaur2023DAHiTrA}. A building-level label identifies which buildings are damaged, but not where on the roof the damage lies or how much of the roof it covers. Damaged roof area and damage ratio require pixel-level information instead: each roof pixel needs an assigned condition category so that damaged pixels can be counted against the roof area.

Studies that localize damage within roofs are fewer and may not satisfy this requirement. For example, the damaged fraction of a roof has been estimated from texture analysis of pre- and post-event images \cite{Radhika2015CycloneDamage} and from superpixel classification inside building outlines taken from GIS data \cite{Boge2020RoofDamage}. Among deep learning approaches, Xu et al. \cite{Xu2022RoofDamage} delineated roof outlines, blue tarps, and completely destroyed roofs and derived damage rates from the tarp-covered and destroyed area, which registers partial damage only where a tarp has been installed. Kucharczyk et al. \cite{Kucharczyk2025RoofDamage} segmented exposed decking and roof holes in imagery of 2 to 15~cm resolution without delineating roofs, so the damaged area cannot be related to the roof area. To the best of the authors' knowledge, no existing method segments the entire roof surface into intact, damaged, covered, and non-evaluable areas at the pixel level from post-event imagery alone, although an area-based damage ratio requires exactly this.

\begin{figure}[!ht]
  \centering
  \includegraphics[width=\textwidth]{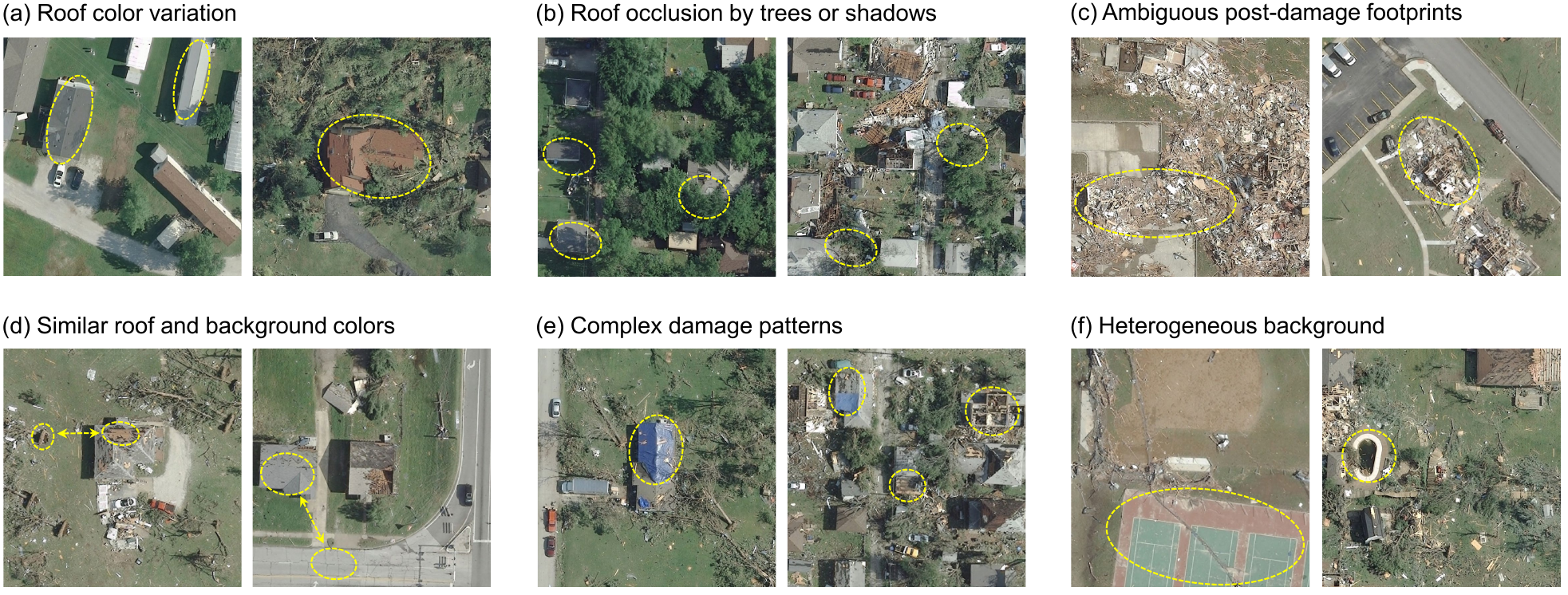}
  \caption{Representative challenges in pixel-level roof damage segmentation from post-wind-hazard aerial imagery: (a) roof color variation, (b) occlusion by trees or shadows, (c) ambiguous post-damage footprints, (d) similar roof and background colors, (e) complex damage patterns, and (f) heterogeneous background. Images are part of the dataset used in this study and are obtained from the 2011 Joplin tornado post-hazard dataset \cite{NOAA2011JoplinImagery}.}
  \label{fig:challenges}
\end{figure}

Pixel-level, category-wise segmentation of roofs from post-event imagery is challenging for the following reasons (Fig.~\ref{fig:challenges}):
\begin{enumerate}
    \item The roof extent needs to be recovered from the damaged scene itself. Pre-event images are often unavailable \cite{AlShafian2025DamageScope}, which rules out change-based methods in many cases \cite{Dong2013BuildingDamageReview,Radhika2015CycloneDamage}, and existing building footprints can be outdated or misaligned and do not reflect destroyed or rebuilt structures \cite{Kucharczyk2025RoofDamage,Xu2022RoofDamage}. However, the post-event image no longer shows the contour of the intact building \cite{Shen2022BDANet}. Severely damaged roofs even lose their edges and merge with the surrounding debris (Fig.~\ref{fig:challenges}c), and roofs can resemble adjacent ground surfaces as well (Fig.~\ref{fig:challenges}(d,f)).
    \item Several conditions coexist within one roof: intact covering, exposed decking or framing, and temporary tarps occupy adjacent areas, vary with roofing material and color (Fig.~\ref{fig:challenges}(a,e)), and can approach the limit of the image resolution \cite{Miura2020BlueTarp,Womble2008WindDamage,Nex2019OperationalDamage}.
    \item Part of the roof may not be evaluated at all. Tree canopy and shadow hide the roof surface (Fig.~\ref{fig:challenges}b) and are a recurrent source of detection errors \cite{Kucharczyk2025RoofDamage,Xu2022RoofDamage}, so these areas need to be identified and excluded from the damage ratio.
\end{enumerate}
A method that can quantify roof damage is desperately needed to resolve all these difficulties at once, from a single post-event image.

This study addresses this challenge with a two-phase Mask2Former framework that operates on post-event aerial imagery alone. The framework takes a task-decomposition approach: Phase I delineates the roof footprint, and Phase II assigns each roof pixel to one of five condition classes that cover intact, partially damaged, severely damaged, tarp-covered, and shadowed roof areas. 
Compared to building-level classification, it resolves damage within each roof. Compared to earlier delineation of roof damage, it requires no external roof outlines or pre-event images, and it classifies the entire roof surface, including areas that cannot be evaluated, instead of mapping damage features or tarps alone. 
The outputs are converted into damaged roof area and a damage ratio for each image tile, with the ratio defined over the evaluable roof area. The framework was trained and tested on aerial imagery of the 2011 Joplin tornado, compared with a single-phase Mask2Former and with CLIPSeg \cite{Lueddecke2022CLIPSeg} and U-Net++ \cite{Zhou2020UNetPlusPlus} variants under the same data partition, and adapted to imagery of Hurricane Melissa (2025). 
On Joplin test dataset, the mean relative deviations from the human-annotated values are 27.69\% for damaged roof area and 6.66\% for damage ratio, lower than those of the three alternatives (29.17--56.89\% and 9.93--13.71\%, respectively). Fine-tuning on Hurricane Melissa images raised the mean intersection over union (mIoU) on held-out images from 0.500 to 0.728. These results demonstrate that roof damage can be quantified, and not only detected, from the aerial imagery collected within days of a wind event, and they indicate that separating roof delineation from condition recognition preserves the roof extent where damage is most severe.

\section{Methodology and dataset}

\subsection{Joplin tornado imagery and annotation}
\label{sec:data}

\begin{figure}[!ht]
  \centering
  \includegraphics[width=\textwidth]{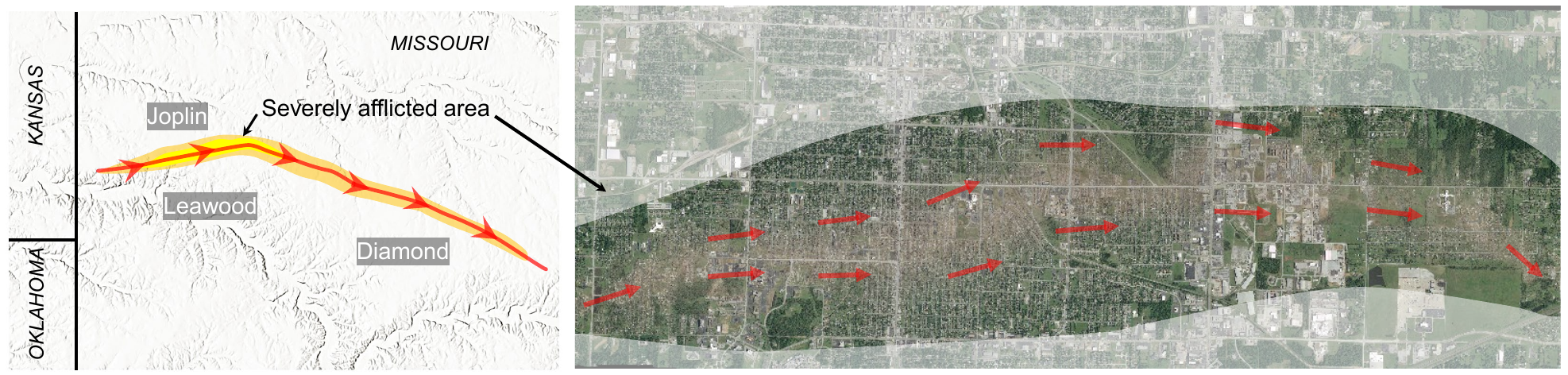}
  \caption{Tornado path, severely affected area, and post-event imagery used to construct the Joplin dataset in this study. The post-event aerial imagery on the right was obtained from NOAA \cite{NOAA2011JoplinImagery}.}
  \label{fig:data}
\end{figure}

On May 22, 2011, an EF5 tornado struck Joplin, Missouri, with estimated maximum wind speeds exceeding 200~mph (322~km/h). Its complete track was approximately 22.1~miles (35.6~km) long and up to $\sim$1~mile (1.6~km) wide. Within Joplin, the EF4--EF5 damage corridor extended $\sim$6~miles (9.7~km) and was generally 0.5--0.75~mile (0.8--1.2~km) wide \cite{NWS2011JoplinTornado}. The event damaged 7,964 residential and nonresidential buildings, including more than 3,000 residences that were heavily damaged or completely destroyed \cite{Kuligowski2014JoplinTornado}. The severely affected part of this geographically defined corridor, where building damage is concentrated, was selected as the study area (Fig.~\ref{fig:data}).

Post-event aerial imagery was obtained from the Remote Sensing Division of the National Oceanic and Atmospheric Administration (NOAA). It was acquired between May 24 and 26, 2011, 2 to 4 days after the tornado, and distributed as orthorectified JPG images with associated world files that map image pixels to map coordinates. The corresponding metadata report a ground sampling distance of 35~cm per pixel \cite{NOAA2011JoplinImagery}. The imagery of the selected study area was partitioned using a non-overlapping regular grid into 182 candidate image tiles of $720 \times 720$ pixels, each covering 252~m by 252~m. The 21 tiles without visible building roofs were excluded, leaving 161 tiles for model development and evaluation. These non-overlapping tiles were randomly divided into mutually exclusive training, validation, and test sets containing 64, 16, and 81 tiles, respectively. The same data partition was used in all subsequent model comparisons.

Roofs were annotated at the pixel level with 5 condition classes (Fig.~\ref{fig:workflow}a). Roof regions that have lost their covering and expose the underlying framing are labeled \emph{damaged} (red). Partially damaged regions, in which patchy or strip-like damage is interspersed with intact roofing, are labeled \emph{minor damage} (magenta). Regions protected by temporary tarps after damage are labeled \emph{covered} (blue). These three classes are damage-related since all of them require roof replacement during repair. They are kept as separate labels to represent their different visual appearance, in line with subclass-based handling of within-class variation \cite{Chen2018SCNN}. Intact regions are labeled \emph{good} (green), and regions whose condition cannot be determined because of shadow or occlusion are labeled \emph{shadowed} (yellow). Pixels outside the annotated roofs constitute the background. The ground-truth (GT) annotations were prepared manually by two annotators and reviewed by a third. They record roof conditions as interpreted from the imagery and may not fully reflect the physical damage to the roof.

\subsection{Two-phase Mask2Former framework}
\label{sec:framework}

\begin{figure}[!ht]
  \centering
  \includegraphics[width=\textwidth]{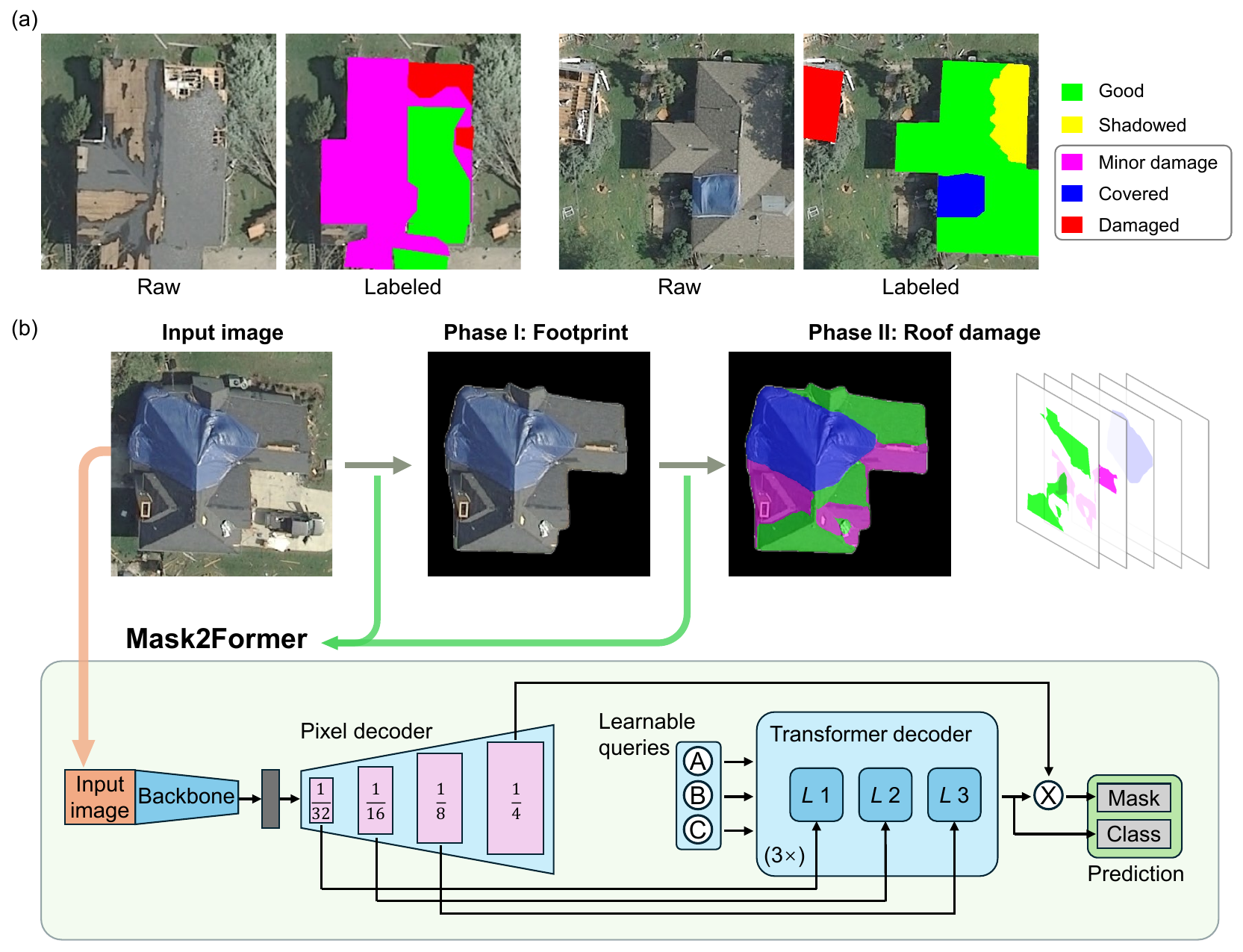}
  \caption{(a) Examples of ground-truth (GT) annotations for the 5 roof condition and damage classes. (b) Mask2Former architecture and the two-phase roof damage prediction workflow.}
  \label{fig:workflow}
\end{figure}

Mask2Former treats segmentation as mask classification: a set of learnable queries is decoded into pairs of a class label and a binary mask \cite{Cheng2021MaskFormer,Cheng2022Mask2Former}, following the query-based set prediction of the DEtection TRansformer (DETR) \cite{Carion2020DETR}. Its architecture combines a feature backbone, a multi-scale pixel decoder, and a Transformer decoder with masked attention, in which each query attends only within its current mask prediction (Fig.~\ref{fig:workflow}b). Mask2Former has been applied to building-footprint extraction and photovoltaic-module segmentation in satellite and aerial imagery \cite{Gibril2024BuildingFootprint, Garcia2024PVSegmentation} and to the segmentation of roof decking and roof holes in post-storm imagery \cite{Kucharczyk2025RoofDamage}. These applications indicate its suitability for delineating spatially structured targets with diverse sizes and appearances in complex overhead scenes. 

The framework assigns roof delineation and condition recognition to two Mask2Former models that share the same architecture, are trained independently, and differ in their output classes. Phase I performs binary segmentation of roof footprint and background, while Phase II assigns pixels to the 5 roof condition classes elaborated in Section~\ref{sec:data} and has no background class. This task decomposition addresses the difficulty of recovering the roof extent in severely damaged scenes. A destroyed roof resembles the debris around it, so a single model that learns roof--background separation together with the condition classes can lose the roof where damage is most severe. In the two-phase design, the footprint model learns from all roof pixels regardless of their condition, and the condition model does not decide what the roof is. The two models can also be adapted separately when the labels available for the two tasks differ, as elaborated in the transfer experiment of Section~\ref{sec:transfer}. The design is also compared with a single six-class model in Section~\ref{sec:single-phase}.

At inference, the Phase I roof-footprint prediction is applied as a spatial mask to the Phase II output (Fig.~\ref{fig:workflow}b). This output-level constraint suppresses class predictions outside the predicted roof footprint before class aggregation and area calculation, which confines the final condition map to the roof. Both models use a Swin-Large Transformer backbone to extract hierarchical feature representations \cite{liu2021swin,shi2026hierarchical}. The multi-scale pixel decoder generates feature maps at \(1/32\), \(1/16\), and \(1/8\) of the input resolution and mask features at \(1/4\) resolution. The Transformer decoder processes the 3 feature scales in coarse-to-fine order and repeats this sequence for 3 rounds, yielding 9 cross-attention layers. The resulting query embeddings are combined with the \(1/4\)-resolution mask features to generate the predicted masks \cite{Cheng2022Mask2Former}. 

\subsection{Model training}
\label{sec:training}

The backbone and Transformer decoder of both phases were initialized from Mask2Former weights pretrained on ADE20K \cite{zhou2017scene}, while the task-specific classification heads were re-initialized for the binary Phase I task and five-class Phase II task. The two phases were optimized independently using the same training and validation sets with the standard Mask2Former objective, which combines Hungarian-matched classification loss with binary cross-entropy and Dice losses on the predicted masks. AdamW \cite{loshchilov2017decoupled,Li2026GranularFlow} was used with a weight decay of 0.01 and learning rates of \(1 \times 10^{-4}\) for the segmentation head and \(1 \times 10^{-5}\) for the pretrained backbone. The learning rate followed a cosine-decay schedule with linear warmup over the first 5\% of the training iterations. Each model was trained for 500 epochs with a batch size of 8 at an input resolution of \(720 \times 720\) pixels. Data augmentation comprised random horizontal and vertical flips and rotations of \(90^\circ\), \(180^\circ\), and \(270^\circ\), applied jointly to the images and labels, and brightness, contrast, and saturation jitter applied to the images only, with each photometric factor sampled from [0.8, 1.2]. For each phase, the checkpoint with the highest validation mean intersection over union (mIoU, Section~\ref{sec:metrics}) over its task-specific label space was selected for inference, with unlabeled pixels excluded.

\subsection{Damage quantification and evaluation metrics}
\label{sec:metrics}

The segmentation outputs were used to calculate roof damage area and roof damage ratio for each image tile. The five Phase II labels were assigned to three aggregated categories over each image-tile domain \(\Omega\): damage-related roof pixels, intact roof pixels, and background/non-evaluated pixels. The damage-related category \(d\) contains the pixels of the class set
\begin{equation}
\mathcal{D}=\{\text{minor damage},\text{covered},\text{damaged}\}.
\end{equation}
The intact category \(g\) contains the pixels labeled \emph{good}. The background/non-evaluated category \(b\) contains the pixels outside the roof footprint and the pixels labeled \emph{shadowed}. Here, ``non-evaluated" refers to exclusion from the area and damage-ratio calculation; these pixels were retained as an aggregated category in the segmentation evaluation based on IoU. 
Shadowed pixels belong to the roof, but their condition cannot be determined visually, so they were counted neither as damaged nor as intact. For model predictions, the aggregation was performed after the Phase II class map had been masked by the Phase I predicted roof footprint. Note that since the 3 damage-related classes are merged, confusion among them, which occurs where their boundaries are visually ambiguous, does not change the damage-related area.

Let \(z_m\in\{d,g,b\}\) denote the category of pixel \(m\). The area of category \(c\) was calculated from the pixel count as
\begin{equation}
A_c=a_p\sum_{m\in\Omega}\mathbb{I}(z_m=c), \qquad c\in\{d,g,b\},
\end{equation}
where \(a_p\) is the pixel-area factor (\(a_p=1\) for the pixel-count-based results reported here) and \(\mathbb{I}(\cdot)\) is the indicator function. For the Joplin imagery, each pixel corresponds to approximately \(0.35\times0.35=0.1225\,\mathrm{m}^2\) in the map plane, and no additional corrections for viewing angle or roof slope were applied in the area calculation. The resulting areas are denoted as \(A_d\), \(A_g\), and \(A_b\). The evaluable roof area and roof damage ratio were defined as:
\begin{equation}
A_t=A_d+A_g,
\end{equation}
\begin{equation}
R_d=\frac{A_d}{A_t}.
\end{equation}
Thus, \(A_t\) includes only roof pixels assigned to the damage-related or intact categories and is not identical to the complete roof footprint, which can also contain shadowed pixels. The background area \(A_b\), including outside-footprint and shadowed pixels, was excluded from the damage-ratio denominator. The same calculation was applied to the GT annotations and to the model predictions, which are distinguished by the subscripts \(\mathrm{GT}\) and \(\mathrm{P}\).

Agreement between the predicted and annotated quantities was measured for each tile by the relative damage-area deviation
\begin{equation}
\label{eq:relative-area-deviation}
D_A=
\begin{cases}
\dfrac{|A_{d,\mathrm{P}}-A_{d,\mathrm{GT}}|}{A_{d,\mathrm{GT}}}\times100\%, & A_{d,\mathrm{GT}}>0,\\[6pt]
\dfrac{|(A_{t,\mathrm{P}}-A_{d,\mathrm{P}})-(A_{t,\mathrm{GT}}-A_{d,\mathrm{GT}})|}{A_{t,\mathrm{GT}}-A_{d,\mathrm{GT}}}\times100\%, & A_{d,\mathrm{GT}}=0,
\end{cases}
\end{equation}
and the relative damage-ratio deviation
\begin{equation}
\label{eq:ratio-deviation}
D_R=
\begin{cases}
\dfrac{|R_{d,\mathrm{P}}-R_{d,\mathrm{GT}}|}{R_{d,\mathrm{GT}}}\times100\%, & R_{d,\mathrm{GT}}>0,\\[6pt]
\dfrac{|(1-R_{d,\mathrm{P}})-(1-R_{d,\mathrm{GT}})|}{1-R_{d,\mathrm{GT}}}\times100\%, & R_{d,\mathrm{GT}}=0.
\end{cases}
\end{equation}
For image tiles with no annotated damage, where \(A_{d,\mathrm{GT}}=0\) and \(R_{d,\mathrm{GT}}=0\), the deviations were calculated from the intact roof area \(A_t-A_d\) and the intact roof ratio \(1-R_d\), respectively. As a special case, a prediction with no evaluable roof area (\(A_{t,\mathrm{P}}=0\)) was treated as a failure, since image tiles without visible roofs had already been excluded from the dataset. To distinguish this failure from a predicted fully damaged roof (\(R_{d,\mathrm{P}}=100\%\)), such a tile was assigned \(R_{d,\mathrm{P}}=0\) and \(D_R=100\%\). These relative deviation indicators summarize disagreement in the image-tile-level quantities derived from pixel-level segmentation \cite{Hussaine2025TunnelSettlement}. Values closer to 0\% indicate closer agreement with the GT annotations.

The area-based deviation indicators describe the quantities derived from the segmentation but not the segmentation itself. Pixel-level agreement was measured by the intersection over union (IoU) \cite{Hussaine2025LandslideSegmentation} of the 3 categories \(d\), \(g\), and \(b\). Note that while the background category \(b\) was excluded from the damage-ratio calculation, it remained an evaluated category in these segmentation metrics. The metrics are computed on the final output, after footprint masking and class aggregation, and differ in this respect from the five-class validation mIoU used to select the Phase II checkpoint. Let \(p_{ij}\) be the number of pixels in \(\Omega\) whose GT category is \(i\) and predicted category is \(j\). The IoU for category \(c\) is:
\begin{equation}
\mathrm{IoU}_c=\frac{p_{cc}}{\sum_{j=1}^{K}p_{cj}+\sum_{i=1}^{K}p_{ic}-p_{cc}},
\end{equation}
where \(K\) is the number of categories present in the GT or predictions: \(K=3\) when all three are present, and \(K=2\) when one category is absent from both. The frequency-weighted IoU (fwIoU) summarizes the segmentation overlap across the complete image tile:
\begin{equation}
\label{eq:fwiou}
\mathrm{fwIoU}=\sum_{c=1}^{K}\frac{n_c}{N_{\mathrm{eval}}}\mathrm{IoU}_c,
\end{equation}
where \(n_c=\sum_{j=1}^{K}p_{cj}\) is the number of GT pixels in category \(c\), and \(N_{\mathrm{eval}}=\sum_{c=1}^{K}n_c\). By weighting category-specific IoU according to GT pixel frequency, fwIoU provides an overall summary of segmentation agreement across the complete image tile that reflects its GT category composition. However, this weighting may conceal weak agreement in a small category when a dominant category is predicted well. Therefore, fwIoU is reported together with mean IoU (mIoU), which gives each category equal weight:
\begin{equation}
\label{eq:miou}
\mathrm{mIoU}=\frac{1}{K}\sum_{c=1}^{K}\mathrm{IoU}_c.
\end{equation}
For pooled values, \(p_{ij}\) was summed over all test tiles. Together, the area and damage-ratio deviation indicators evaluate agreement in the image-tile-level quantities, while mIoU and fwIoU jointly provide complementary summaries of the pixel-level segmentation quality underlying those quantities.

\section{Results for Joplin tornado}

\subsection{Representative cases under diverse roof-damage conditions}
\label{sec:cases}

\begin{figure}[!h]
  \centering
  \includegraphics[width=\textwidth]{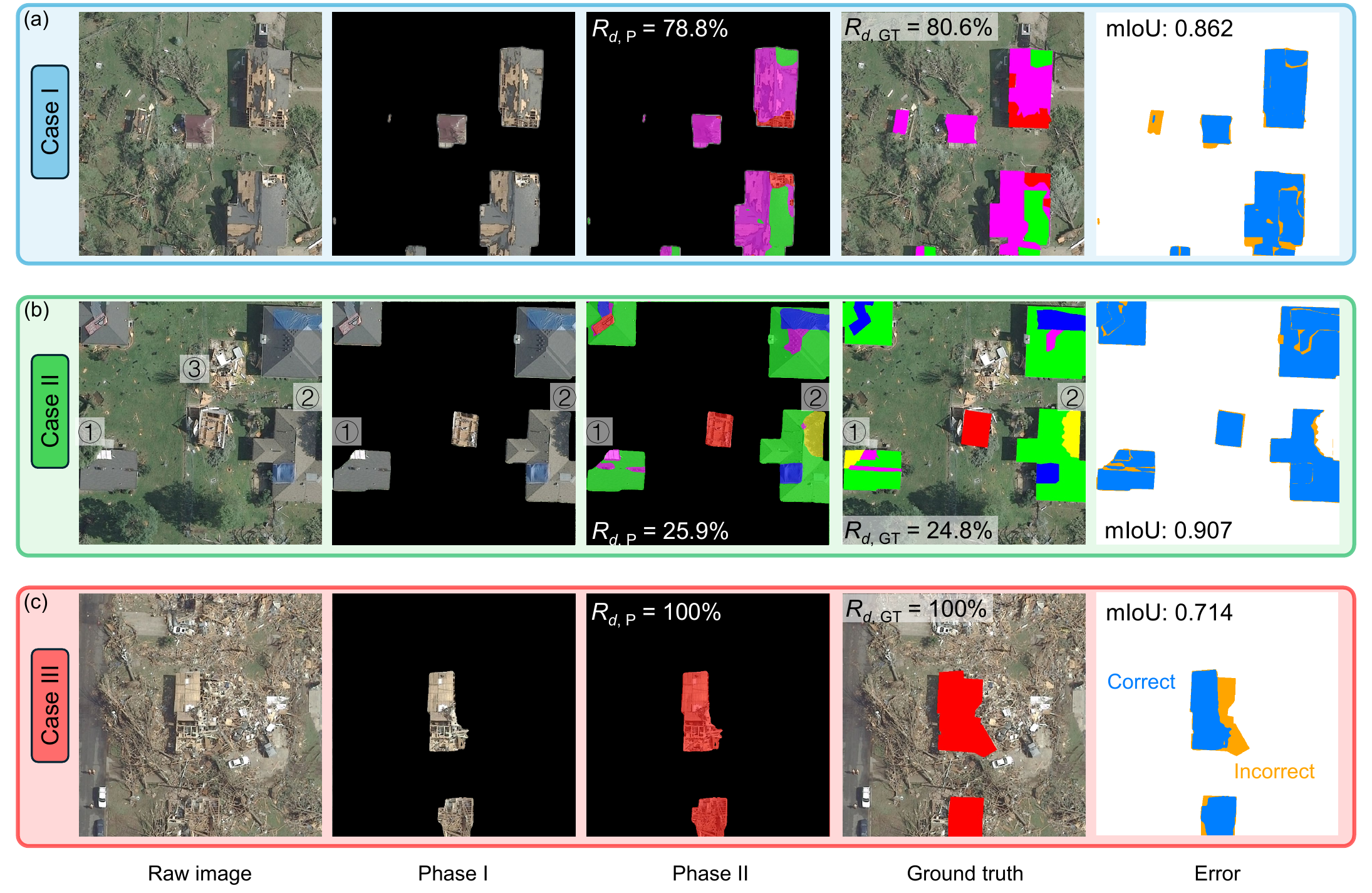}
  \caption{Representative results of the framework for 3 Joplin image tiles: (a) partial damage with relatively clear roof boundaries (Case I), (b) mixed roof conditions with tree cover and shadow (Case II), and (c) complete damage with roof--debris boundary ambiguity (Case III). Columns show the raw image, Phase I roof-footprint prediction, Phase II condition (damage) prediction, ground-truth (GT) annotation, and segmentation error map.}
  \label{fig:main-results}
\end{figure}

Post-event roof imagery contains diverse combinations of damage extent, roof geometry, visibility obstruction, and roof--background ambiguity. Three Joplin image tiles were selected to represent different demands on the framework (Fig.~\ref{fig:main-results}): partial damage on roofs where intermediate damage needs to be recovered from a mixture of intact and damaged pixels (Case~I), mixed and complex roof conditions with tree cover and shadow (Case~II), and complete visible damage with an ambiguous roof--debris boundary (Case~III). Together, these cases illustrate the importance of area normalization, visibility handling, and roof-footprint delineation on the resulting tile-level roof damage estimates.

In Case~I (Fig.~\ref{fig:main-results}a), the roof boundaries are relatively clear, and the task reduces to separating intact from damaged regions on the same roofs. Phase I retains the principal roof footprints, while Phase II distinguishes intact and damage-related pixels within them. The resulting roof damage ratio, \(R_{d,\mathrm{P}}=78.8\%\), is close to \(R_{d,\mathrm{GT}}=80.6\%\), corresponding to \(D_R=2.2\%\) and \(D_A=3.8\%\). This case illustrates how the two-phase workflow converts spatially heterogeneous damage masks into consistent tile-level estimates when the roof boundaries remain visually identifiable.

Case II (Fig.~\ref{fig:main-results}b) demonstrates how visibility-related uncertainty was handled in our framework. The image tile contains intact, covered, damaged, and tree-obstructed roof regions. At \Circled{1}, tree cover hides the roof outline, and the region is treated as background. At \Circled{2}, the roof footprint remains identifiable, but the surface condition is obscured by tree shadow, and the region is assigned to the \emph{shadowed} class. Both regions are excluded from \(A_t\) and \(A_d\), and the excluded shadowed/occluded area accounts for 6.9\% of the annotated roof extent before exclusion. This explicit exclusion keeps visually indeterminate pixels out of both the intact and the damage-related category. At \Circled{3}, a region that stands out from its surroundings is treated as background by the framework, and all three annotators confirmed this prediction as the region is debris rather than a roof. The predicted ratio, \(R_{d,\mathrm{P}}=25.9\%\), remains close to \(R_{d,\mathrm{GT}}=24.8\%\), with \(D_R=4.5\%\) and \(D_A=5.1\%\), illustrating how visibility-aware segmentation provides a responsible damage quantification by preserving a consistent evaluable-area definition in scenes containing mixed roof conditions and obstruction.

In Case III (Fig.~\ref{fig:main-results}c), the completely damaged roof regions and the adjacent debris have similar visual textures, making their boundary difficult to delineate consistently even during manual annotation. The predicted damage area is \(A_{d,\mathrm{P}}=31{,}365\) pixels, compared with \(A_{d,\mathrm{GT}}=43{,}076\) pixels, yielding \(D_A=27.2\%\). Nevertheless, both the predicted and annotated roof damage ratios are 100\% because all evaluated roof pixels are classified as damaged. In this case, the damage ratio retains the relative damage extent when boundary differences affect the numerator and denominator in the same direction, while the damage area reveals the remaining disagreement in footprint delineation. The two quantities provide complementary information rather than allowing agreement in a single normalized metric to obscure spatial error.

The three cases span a broad range of roof-damage assessment conditions, from partial to complete damage and from clearly identifiable roof regions to mixed, obscured, or debris-adjacent surfaces. Across these contrasting scenarios, the developed framework produces spatial damage maps and corresponding damage-area and damage-ratio estimates, supporting its applicability under varied post-event roof conditions. This breadth of damage conditions demonstrates the engineering applicability for rapid hazard reconnaissance, where the assessment needs to account for the variability in both damage extent and roof visibility.


\subsection{Overall test-set performance}

\begin{figure}[!ht]
  \centering
  \includegraphics[width=\textwidth]{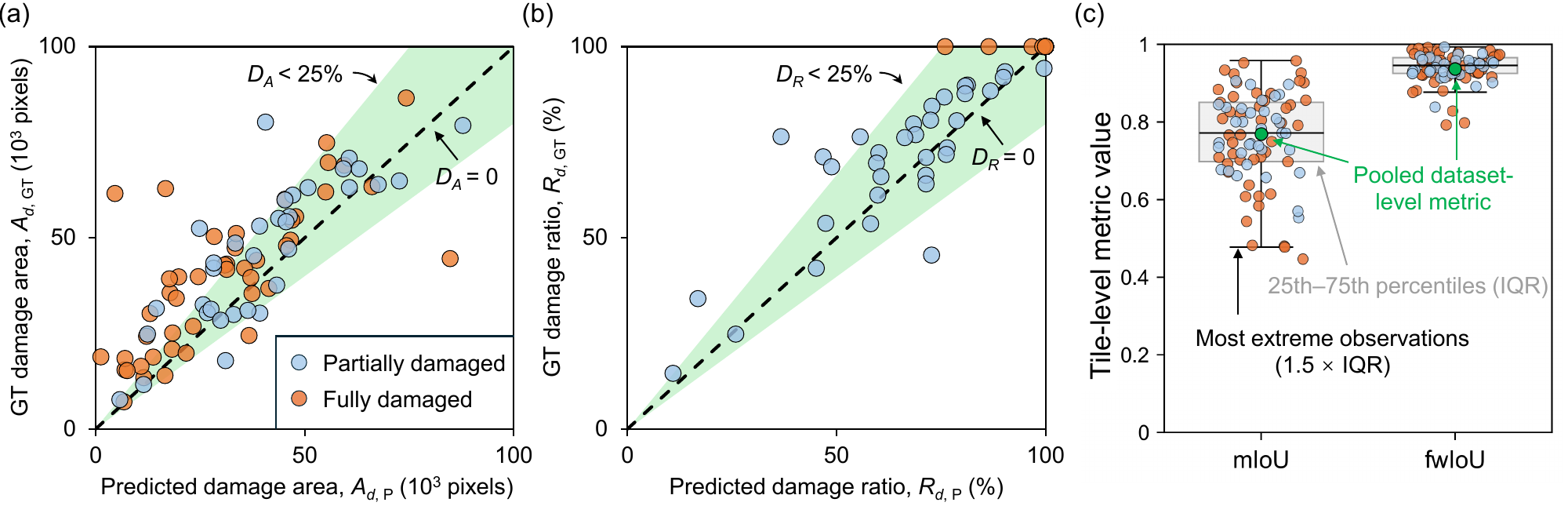}
  \caption{Predicted and GT values for (a) damage area and (b) roof damage ratio, with the regions \(D_A<25\%\) and \(D_R<25\%\) shaded. (c) Distributions of tile-level mIoU and fwIoU, with the pooled metrics over the test set highlighted.}  
  \label{fig:full-results}
\end{figure}

The test set comprised 81 tiles, including 47 with \(R_{d,\mathrm{GT}}=100\%\) (fully damaged) and 34 with \(0<R_{d,\mathrm{GT}}<100\%\) (partially damaged). Phase I achieved a high test-set mIoU of 0.963 for the binary roof-footprint segmentation task. The following analysis focuses on the final predictions of the framework. Fig.~\ref{fig:full-results} compares the predicted and GT damage quantities and summarizes the tile-level segmentation metrics. The shaded regions \(D_A<25\%\) and \(D_R<25\%\) serve as descriptive reference bands.

Damage-area prediction was evaluated using the relative damage-area deviation \(D_A\) defined in Eq.~\eqref{eq:relative-area-deviation}, with smaller values indicating closer agreement between the predicted and GT damage areas, \(A_{d,\mathrm{P}}\) and \(A_{d,\mathrm{GT}}\). The median and mean \(D_A\) were 20.81\% and 27.69\%, respectively, and 47 of the 81 tiles (58.0\%) fell within the 25\% band (Fig.~\ref{fig:full-results}a). Among the 34 out-of-band tiles, 24 (70.6\%) were fully damaged, and most of these lie above the 1:1 line, where the predicted damage area is smaller than the annotated one. As illustrated by Case III in Fig.~\ref{fig:main-results}c, extensive roof damage and surrounding debris can make the roof footprint boundary difficult to distinguish even during manual annotation. Therefore, an elevated \(D_A\) under such conditions is an interpretable boundary-related failure mode, and \(D_A\) should be considered jointly with \(D_R\) when evaluating these tiles. By contrast, within the partially damaged group, 24 of 34 tiles (70.6\% of that group) fell within the band, indicating close damage-area agreement for most tiles in this group. This pattern is consistent with Case I, in which visually identifiable roof boundaries coincided with a low \(D_A\).

For the relative roof damage-ratio deviation \(D_R\) (Eq.~\eqref{eq:ratio-deviation}), the median and mean were 0\% and 6.66\%, respectively (Fig.~\ref{fig:full-results}b). The zero median reflects the large number of exact damage-ratio predictions in the fully damaged group: 43 of the 47 tiles with \(R_{d,\mathrm{GT}}=100\%\) also yielded \(R_{d,\mathrm{P}}=100\%\), resulting in \(D_R=0\%\) for these tiles. For the partially damaged tiles, the median and mean \(D_R\) were 10.06\% and 14.64\%, respectively, showing that close damage-ratio agreement is not limited to the fully damaged group. In total, 74 of the 81 tiles (91.4\%) fell within the 25\% band. The ratio deviation \(D_R\) is less sensitive than \(D_A\) because a difference in the footprint boundary affects \(A_d\) and \(A_t\) in the same direction. The results indicate that using the two indicators together reveals discrepancies that either one alone would hide \cite{Shi2026SoilPlugAgents}.

The tile-level mIoU (Eq.~\eqref{eq:miou}) had a median of 0.772, with the 25th and 75th percentiles at 0.698 and 0.851 (Fig.~\ref{fig:full-results}c), and the pooled mIoU was 0.770. Building-damage segmentation in post-hazard imagery is often challenged by small damage regions, imbalanced class distributions, ambiguous boundaries, occlusion, and complex background textures, all of which can reduce category-specific IoU \cite{Xiong2022TornadoDamage,Wang2025HistoricDamage}. Nevertheless, the pooled mIoU (0.770) represents acceptable performance for this application domain \cite{Wu2025PostFloodDamage}. Compared with mIoU, fwIoU had a higher median of 0.946 and a smaller interquartile range, from 0.925 to 0.966, with a pooled dataset-level fwIoU of 0.937. The gap between the two metrics follows from their weighting (Section~\ref{sec:metrics}): the fwIoU gives most weight to the categories that occupy most of the tile, while the mIoU weights the three categories equally, so poor prediction in a category can lower the overall score even when that category occupies only a small part of the image. Used together, the metrics retain sensitivity to poorly predicted categories while also accounting for their relative area in the image. When interpreted alongside \(D_A\) and \(D_R\), the two metrics provide pixel-level segmentation information that complements the evaluation of damage area and damage ratio.

\section{Comparative analyses and discussion}

This section examines the two design choices of the framework: the separation into two phases (Section~\ref{sec:single-phase}) and the use of Mask2Former for condition segmentation (Sections~\ref{sec:phase2-models} and~\ref{sec:interpretation}). It then tests the transfer of the framework to imagery from a different hazard (Section~\ref{sec:transfer}). 
Table~\ref{tab:comparison} collects the test-set indicators of the 4 configurations compared on the Joplin data.

\begin{table}[!ht]
  \centering
  \caption{Test-set evaluation metrics of the 4 configurations compared on the Joplin image tiles.}
  \label{tab:comparison}
  \begin{tabular}{lcccc}
    \toprule
    & Mean & Mean & Median & Median \\
    Configuration & \(D_A\) (\%) & \(D_R\) (\%) & mIoU & fwIoU \\
    \midrule
    Two-phase framework (Mask2Former) & 27.69 & 6.66 & 0.772 & 0.946 \\
    Single-phase baseline (Mask2Former) & 47.14 & 11.07 & 0.702 & 0.922 \\
    Two-phase variant, CLIPSeg in Phase II & 29.17 & 9.93 & 0.704 & 0.940 \\
    Two-phase variant, U-Net++ in Phase II & 56.89 & 13.71 & 0.569 & 0.898 \\
    \bottomrule
  \end{tabular}
\end{table}

\subsection{Two-phase versus single-phase segmentation}
\label{sec:single-phase}

The separation of footprint delineation from condition segmentation was tested against a single-phase baseline: one Mask2Former trained with the background as a 6th class next to the 5 roof condition classes, so that roof--background separation and condition labeling are learned jointly. Except for this task formulation and the six-class output, the single-phase baseline used the same network architecture, data partition, and per-model training and checkpoint-selection settings as the two-phase framework. Its output was aggregated into the same three categories and evaluated using the same metrics (Section~\ref{sec:metrics}).

\begin{figure}[!ht]
  \centering
  \includegraphics[width=\textwidth]{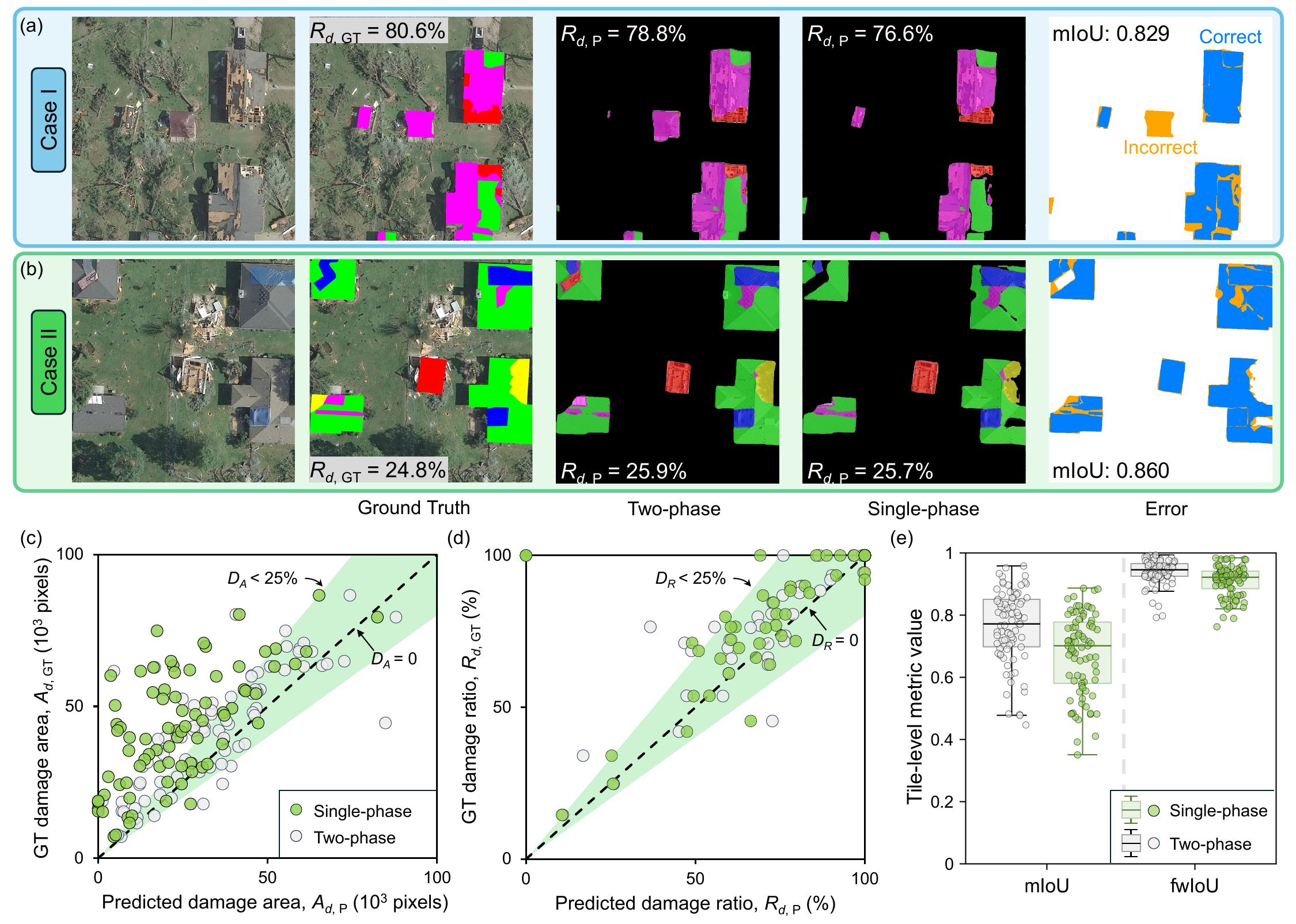}
  \caption{Comparison between the two-phase segmentation framework and the corresponding single-phase baseline. (a, b) Raw images, GT annotations, predictions of both approaches, and error maps of the single-phase baseline for Cases~I and II. (c) Predicted and GT roof damage areas \(A_{d,\mathrm{P}}\) and \(A_{d,\mathrm{GT}}\). (d) Predicted and GT roof damage ratios \(R_{d,\mathrm{P}}\) and \(R_{d,\mathrm{GT}}\). (e) Distributions of tile-level mIoU and fwIoU. The bands \(D_A<25\%\) and \(D_R<25\%\) are shaded in (c) and (d).}
  \label{fig:twophase-singlestage}
\end{figure}

Fig.~\ref{fig:twophase-singlestage}(a,b) compares the two approaches for Cases~I and II introduced in Section~\ref{sec:cases}. In both cases, the single-phase estimated damage ratios (\(R_{d,\mathrm{P}}=76.6\%\) and 25.7\%) lie within 5\% of the GT values of 80.6\% and 24.8\%, as do the 78.8\% and 25.9\% predicted by the proposed two-phase framework. However, compared with the two-phase predictions, the single-phase masks contain clear interruptions, i.e., pixels within the annotated roof regions that were assigned to background (shown as black pixels in the masked tiles). Such errors can leave \(R_d\) numerically close to the GT value while omitting damaged roof pixels and their locations from the spatial output, which are important for spatially resolved post-event damage assessment. The single-phase algorithm also entirely missed one annotated roof in Case~I, while the developed two-phase framework retained it. Together, these comparisons demonstrate the necessity of having roof-footprint segmentation (Phase I) before the condition segmentation, especially when severe damage weakens roof--background boundary hints.

The same pattern holds across the test set (Fig.~\ref{fig:twophase-singlestage}c--e and Table~\ref{tab:comparison}). The single-phase baseline had a mean relative damage-area deviation \(D_A\) of 47.14\%, which was 70.24\% higher than that of the two-phase framework (27.69\%). More of the single-phase points lay above the 1:1 line in Fig.~\ref{fig:twophase-singlestage}c, indicating more frequent underestimation of the damage area \(A_d\). As shown in Fig.~\ref{fig:twophase-singlestage}d, the mean relative damage-ratio deviation \(D_R\) was 6.66\% for the two-phase model and 11.07\% for the single-phase baseline. The smaller gap for \(D_R\) may partly reflect the simultaneous underestimation of damage area and evaluable roof area by the single-phase baseline, allowing the two errors to partially cancel in the normalized ratio. Moreover, several single-phase observations lie at \(R_{d,\mathrm{GT}}=100\%\) and \(R_{d,\mathrm{P}}=0\): in these fully damaged tiles the baseline found no roof area, and the tiles were assigned \(D_R=100\%\) (Section~\ref{sec:metrics}). This is the failure anticipated in Section~\ref{sec:framework}, and it is the most consequential one for post-event screening, since a tile with complete roof damage then appears to contain no roof. Fig.~\ref{fig:twophase-singlestage}e provides a complementary pixel-level comparison. The median mIoU and fwIoU of the baseline were 0.702 and 0.922, below the 0.772 and 0.946 of the proposed framework, and both of its distributions were wider. Taken together, the higher medians and tighter distributions due to the dedicated footprint model in the proposed two-phase framework indicate greater and more consistent pixel-level agreement across the test set, providing a more stable basis for deriving roof damage areas and damage ratios.

\subsection{Comparison with alternative segmentation models}
\label{sec:phase2-models}

The choice of Mask2Former for damage condition segmentation was examined by replacing the Phase II model with U-Net++ and with CLIPSeg. U-Net++ is a convolutional encoder--decoder architecture that integrates multi-scale features through nested, densely connected skip pathways \cite{Zhou2020UNetPlusPlus,iakubovskii2019segmentation}. CLIPSeg extends CLIP with a lightweight prompt-conditioned decoder to generate segmentation masks from natural-language queries \cite{radford2021learning,Lueddecke2022CLIPSeg}. Both of them have publicly available implementations and pretrained weights, and they represent convolutional and vision-language alternatives to the query-based Mask2Former.

To facilitate a direct comparison, the trained Phase I model was kept, its roof footprint was applied to the output of every Phase II model as the same spatial mask, and only the Phase II segmentation model was varied. Both baseline variants used the same Phase II five-class label space and the test set of the proposed framework. In contrast to Mask2Former, which predicts class--mask pairs through query-based masked attention, U-Net++ generated the five-class segmentation map by fusing convolutional features across the encoder and decoder scales and was trained using an equally weighted combination of cross-entropy and Dice losses. This configuration emphasizes convolutional multi-scale feature fusion and direct pixel-overlap optimization, which may affect the delineation of spatially fragmented damage regions. CLIPSeg followed a different strategy by treating each roof class as an independent text-prompted binary segmentation task and combining the 5 resulting masks according to the highest response at each pixel. Since the masks were predicted independently, the final class assignment depended on their relative response scores. Moreover, the CLIP image and text encoders were frozen during training \cite{Fu2026IAP}, and only the segmentation decoder was updated \cite{amini2025debris}. Therefore, CLIPSeg relied primarily on its general pretrained image--text representations, while Mask2Former adapted both its feature extractor and decoder to the roof-damage data; this difference may limit CLIPSeg's ability to distinguish roof-condition classes with similar visual characteristics. Both baseline variants were trained for 300 epochs, and the checkpoint with the highest validation mIoU was retained for each model, consistent with the Mask2Former framework.

\begin{figure}[!ht]
  \centering
  \includegraphics[width=\textwidth]{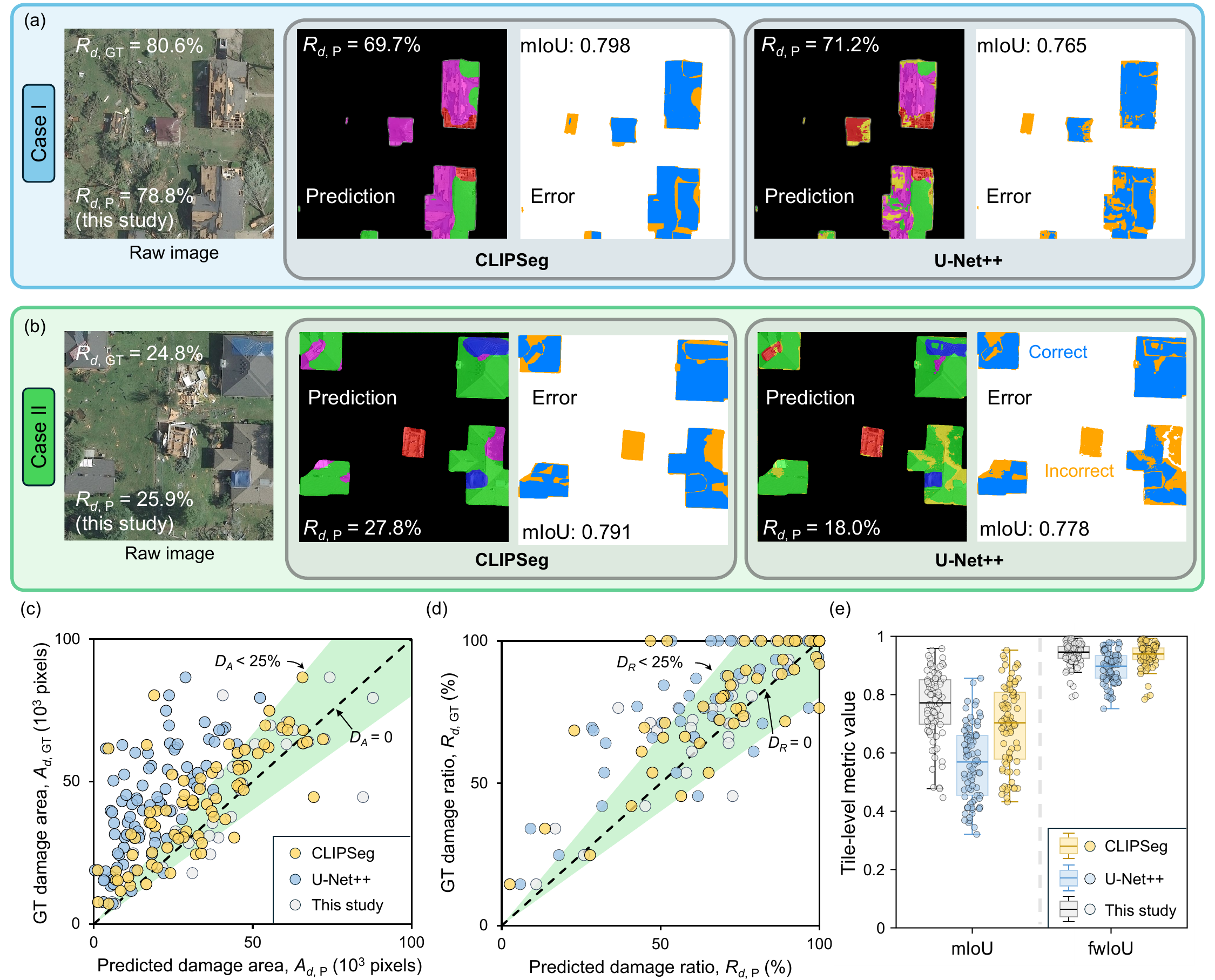}
  \caption{Comparison of the proposed two-phase Mask2Former framework against CLIPSeg- and U-Net++-based variants. (a, b) Raw images, predictions, and error maps for Cases~I and II from the two variants. (c) Predicted and GT roof damage areas \(A_{d,\mathrm{P}}\) and \(A_{d,\mathrm{GT}}\). (d) Predicted and GT roof damage ratios \(R_{d,\mathrm{P}}\) and \(R_{d,\mathrm{GT}}\). (e) Tile-level mIoU and fwIoU distributions. The shaded regions in panels (c) and (d) denote the descriptive \(D_A<25\%\) and \(D_R<25\%\) reference bands, respectively.}
  \label{fig:baseline}
\end{figure}

Fig.~\ref{fig:baseline}(a,b) presents the CLIPSeg- and U-Net++-based predictions for Cases~I and II. For Case~I, a partial-damage image tile with relatively clear roof boundaries, CLIPSeg and U-Net++ predicted \(R_{d,\mathrm{P}}=69.7\%\) and 71.2\% against \(R_{d,\mathrm{GT}}=80.6\%\). Their corresponding relative damage-ratio deviation \(D_R\) values were 13.5\% and 11.6\%, $\sim 6.1$ and 5.2 times the two-phase Mask2Former prediction of 2.2\%, respectively. For Case~II, which contains mixed damage conditions together with tree cover and shadow, CLIPSeg and U-Net++ predicted \(R_{d,\mathrm{P}}=27.8\%\) and 18.0\%, respectively, while \(R_{d,\mathrm{GT}}=24.8\%\). The resulting \(D_R=12.3\%\) and 27.2\%, $\sim 2.7$ and 6.1 times the Mask2Former prediction of 4.5\%. The tile-level mIoU is also highest for the proposed framework: 0.868 in Case~I, against 0.798 for CLIPSeg and 0.765 for U-Net++, and 0.906 in Case~II, against 0.791 and 0.778 (Figs.~\ref{fig:main-results} and~\ref{fig:baseline}). Thus, the numerical results alone show a clear advantage of Mask2Former in these selected cases, as it produced the smallest damage-ratio deviation in both cases.

The pixel-level error maps further reveal the weaknesses of the two baseline models. For CLIPSeg, the principal damaged and intact regions in Case~I were generally identified correctly, so the overall class assignment was reasonable; the errors were instead concentrated along the interfaces between them. In Case~II, CLIPSeg failed to identify the annotated shadowed regions and assigned most of them to a damage-related class. These errors may arise because CLIPSeg predicts each class independently using predominantly frozen image--text representations, which limits both the separation of visually similar roof conditions and the localization of their boundaries. U-Net++ exhibited a different and more extensive weakness. Although its \(D_R\) was lower than that of CLIPSeg in Case~I, its prediction was spatially fragmented, with frequent pixel-scale transitions among intact, damage-related, and shadowed classes. Similar fragmentation was observed in Case~II, where shadow predictions were also distributed across regions annotated as other roof conditions. This pattern may reflect the sensitivity of dense pixelwise classification to locally ambiguous features: unlike Mask2Former, U-Net++ does not explicitly group pixels into class-labeled region masks, allowing neighboring pixels with similar visual characteristics to switch repeatedly between classes. Such locally unstable predictions reduce the reliability and interpretability of the resulting damage map as a basis for spatially resolved engineering assessment.

Over the test set (Fig.~\ref{fig:baseline}c--e and Table~\ref{tab:comparison}), CLIPSeg stayed close to the proposed framework in the damage area, with a mean relative damage-area deviation \(D_A\) of 29.17\% against 27.69\%. U-Net++ reached 56.89\%, and most of its points lie above the 1:1 line, which indicates frequent underestimation of the damage area. The mean relative damage-ratio deviation \(D_R\) was 9.93\% for CLIPSeg and 13.71\% for U-Net++, compared with 6.66\% for the proposed framework (Fig.~\ref{fig:baseline}d). For CLIPSeg, the larger \(D_R\) may partly reflect its poor recognition of shadowed regions, which moves non-evaluable pixels into the damage-related category and raises the damage ratio. Fig.~\ref{fig:baseline}e further compares the models using pixel-level segmentation metrics. CLIPSeg and U-Net++ achieved median mIoU of 0.704 and 0.569, respectively, which were 8.83\% and 26.26\% lower than the proposed framework. Their median fwIoU values were 0.940 and 0.898, corresponding to reductions of 0.70\% and 5.12\%, respectively. Together with the \(D_A\) and \(D_R\) comparison, such results demonstrate that Mask2Former provides the lowest deviations and the highest pixel-level agreement among the 3 Phase II models evaluated.

\subsection{Query-level interpretation of the Phase II model}
\label{sec:interpretation}

\begin{figure}[!ht]
  \centering
  \includegraphics[width=\textwidth]{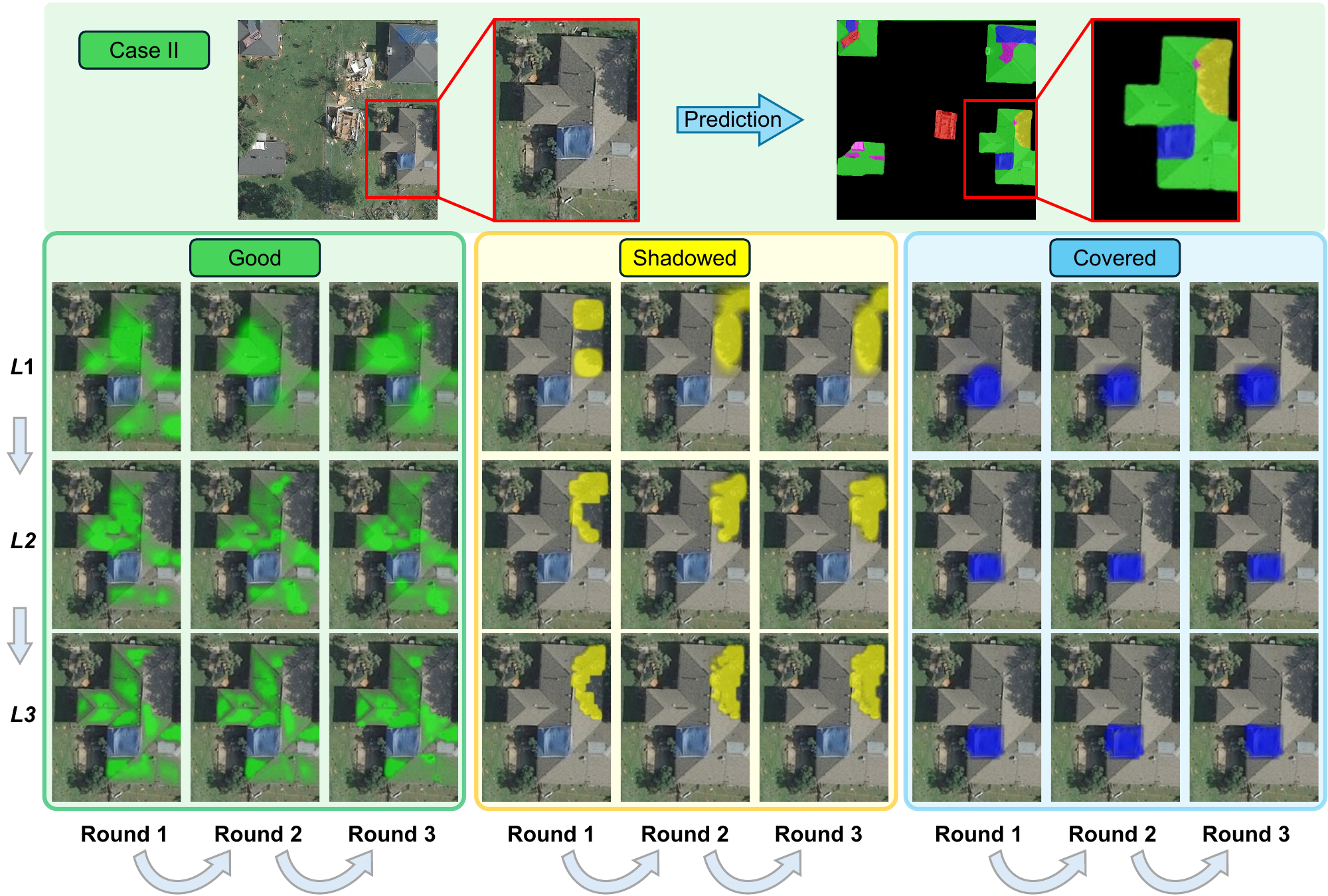}
  \caption{Cross-attention maps of the \emph{good}, \emph{shadowed}, and \emph{covered} class queries for a roof of Case~II with 3 adjacent conditions. The upper panels locate the roof in the tile and present its enlarged image and final prediction. The lower panels show the maps at 3 decoder layers (L1--L3) in each of 3 refinement rounds.}
  \label{fig:interpretation}
\end{figure}

In the comparisons above, Mask2Former separated neighboring different roof conditions more cleanly than the two alternatives. To examine this behavior, the cross-attention of the Phase II model was visualized for the roof in the lower-right part of Case~II, which contains adjacent \emph{good}, \emph{shadowed}, and \emph{covered} regions (Fig.~\ref{fig:interpretation}) \cite{hussaine2026geoslideformer}. For each of the 3 classes, the query with the highest predicted probability for that class in the final prediction was followed through the 9 cross-attention layers of the Transformer decoder. Within each decoder round, layers L1--L3 sequentially process the feature maps at \(1/32\), \(1/16\), and \(1/8\) of the input resolution, and this coarse-to-fine sequence is repeated over 3 rounds (Section~\ref{sec:framework}). The panels show the iterative query refinement within a single inference pass rather than stages of model training. Since each masked-attention layer uses the preceding mask prediction to restrict its spatial support, initially diffuse responses can be progressively concentrated around the region represented by each query.

The 3 class queries exhibit distinct refinement trajectories. The \emph{covered} query localizes earliest: its attention is already centered on the tarp-covered area in Round~1 and remains compact through the later layers and rounds. The \emph{shadowed} query initially produces two coarse responses along the right side of the roof, which subsequently merge into a more continuous band corresponding to the shadowed region. By contrast, the \emph{good} query initially spans a broader area because visibly intact roofing extends across several roof planes surrounding the \emph{covered} and \emph{shadowed} regions. As the decoder proceeds to finer-scale layers, its attention increasingly aligns with these intact roof planes and becomes more clearly separated from the adjacent \emph{covered} and \emph{shadowed} regions. These trajectories elaborate how Mask2Former progressively develops localized, class-specific responses with distinct spatial boundaries, even when different roof conditions occur in close proximity.

These maps are qualitative evidence from one roof and are consistent with the comparisons in Section~\ref{sec:phase2-models}, in which Mask2Former produced less fragmented masks than U-Net++ and clearer boundaries between classes than CLIPSeg. Together with Section~\ref{sec:single-phase}, they suggest that the two design choices are complementary: the dedicated footprint model provides roof-level spatial support for the final condition map, while the class queries resolve adjacent roof conditions into spatially distinct regions.

\subsection{Cross-domain transfer and adaptation to Hurricane Melissa imagery}
\label{sec:transfer}

To examine cross-domain performance and supervised adaptation, the framework trained on the Joplin dataset was applied to an independent post-hazard imagery dataset and was evaluated before and after target-domain fine-tuning \cite{Fu2025IncrementalReview}. The target dataset comprised 140 annotated image tiles extracted from NOAA post-event aerial imagery acquired over Jamaica between October 31 and November 6, 2025, after Hurricane Melissa \cite{NOAA2025MelissaImagery}. Unlike tornadoes, hurricanes generally expose buildings to longer-duration winds and wind-driven rain, creating a different damage context. In addition, residential buildings in the target imagery commonly had tile-roof surfaces adjoining paved platforms or terraces due to different building codes and construction protocols. Together, these hazard and architectural differences created roof--background configurations that were absent from the Joplin training data and were expected to limit the direct-transfer performance of the original framework \cite{shi2026artificial}. Since the two phases are trained independently, they could be fine-tuned with different numbers of target tiles: 95 for Phase I and 35 for Phase II, with data augmentation and without any change to the framework architecture. The 35 Phase II tiles are a subset of the 95 Phase I tiles; 30\% of each set was reserved for validation, and both sets were disjoint from the 45 held-out test tiles. For each phase, the checkpoint with the highest validation mIoU over its label space was retained, as for the Joplin models.

\begin{figure}[!ht]
  \centering
  \includegraphics[width=\textwidth]{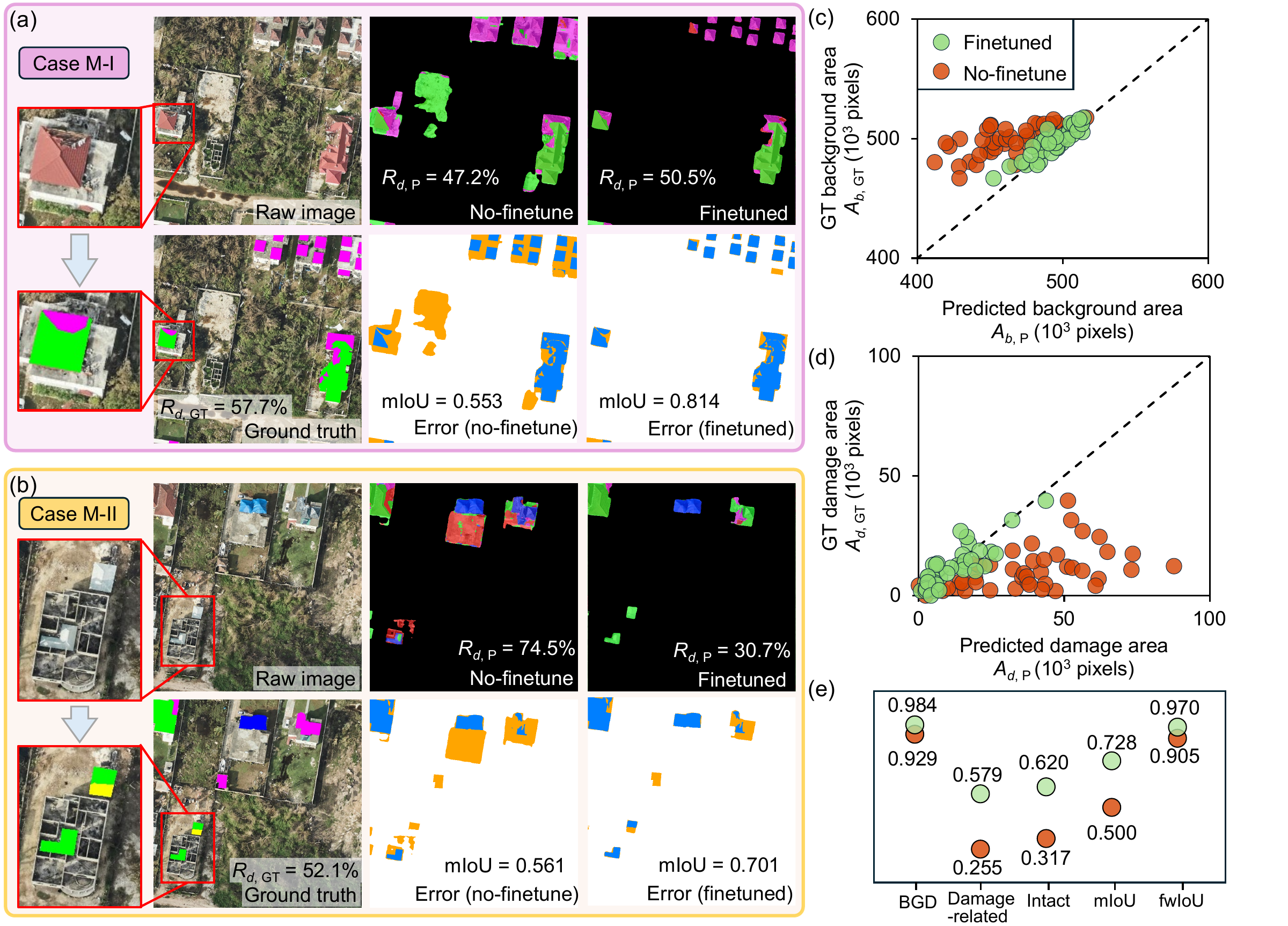}
  \caption{Cross-domain evaluation of the Joplin-trained framework on post-Hurricane Melissa aerial image tiles from Jamaica, before and after target-domain fine-tuning. (a) Case M-I and (b) Case M-II, with raw images, GT annotations, predictions, and error maps. (c) Predicted versus GT area of the background/non-evaluated category. (d) Predicted versus GT damage area for the 45 test tiles, with dashed lines representing perfect agreement. (e) Pooled IoU of the background/non-evaluated (BGD), damage-related, and intact categories, with the pooled mIoU and fwIoU before and after fine-tuning.}
  \label{fig:domain-shift}
\end{figure}

Fig.~\ref{fig:domain-shift}(a,b) compares the predictions for two Hurricane Melissa cases obtained using the proposed framework before and after fine-tuning. In both cases, the original model (before fine-tuning) incorrectly includes large adjoining platforms, terraces, or flat exposed surfaces in the predicted roof footprint, which is expected for the domain shift due to the lack of such training data in the original Joplin imagery. After fine-tuning, these false-positive regions are substantially reduced, and the predicted masks more closely delineate the annotated roof footprints. In Case M-I, the prediction before fine-tuning spreads from the red tile roof onto the adjacent light-colored platform, and after fine-tuning it stays on the roof planes. In Case M-II, fine-tuning suppresses the large false-positive region over the exposed flat surface while retaining the smaller annotated roof regions. Accordingly, fine-tuning increased the tile-level mIoU from 0.553 to 0.814 for Case M-I and from 0.561 to 0.701 for Case M-II, corresponding to relative increases of 47.1\% and 24.9\%, respectively. The predicted damage ratio of Case~M-I moved from 47.2\% to 50.5\%, closer to the annotated 57.7\%. That of Case~M-II moved from 74.5\% to 30.7\% against an annotated 52.1\%, so the ratio remained far from the GT value although the footprint was corrected.

Fig.~\ref{fig:domain-shift}c shows that, before fine-tuning, the predicted background area was lower than the GT value for all 45 test tiles, indicating systematic overextension of the predicted roof footprints into adjacent non-roof regions. After fine-tuning, the prediction points clustered closely around the 1:1 line, and the mean absolute background-area error decreased from 34,290 to 4,990 pixels, representing an 85.5\% reduction. A similar improvement can be observed in Fig.~\ref{fig:domain-shift}d, where the predicted damage-related areas moved substantially closer to their GT values; the corresponding mean absolute error decreased from 26,400 to 3,430 pixels, an 87.0\% reduction. Nevertheless, damage-area predictions for some Melissa tiles remained less accurate than those for Joplin, as the damage ratio of Case~M-II illustrates (Fig.~\ref{fig:domain-shift}b). This lower accuracy largely reflects the small damage regions in many Melissa tiles, which are more difficult to delineate reliably.

The pooled metrics in Fig.~\ref{fig:domain-shift}e confirm the overall improvement for all three categories: the IoU rose from 0.929 to 0.984 for the background/non-evaluated category, from 0.255 to 0.579 for the damage-related category, and from 0.317 to 0.620 for the intact category. Consequently, the pooled mIoU increased from 0.500 to 0.728, and the pooled fwIoU rose from 0.905 to 0.970. Overall, these gains support the cross-domain adaptability of the proposed two-phase transformer-based framework: supervised fine-tuning adapted the framework to different building characteristics and a different damage context without architectural modification.

\section{Conclusions}

This study developed and trained a two-phase Mask2Former framework using post-wind-hazard aerial imagery to segment and quantify roof damage, measured by damaged roof area and the damage ratio (the damaged share within the whole roof area). Phase I delineates roof footprints from the background, and Phase II assigns pixels within the roof footprints to 5 roof condition classes, distinguishing damaged, intact, and background regions. For the 2011 Joplin tornado case study, the predicted damage ratio was within 6.66\% of the manually annotated value on average, and 43 of the 47 image tiles with fully damaged roofs were recognized. The ratio held even where destroyed roofs merge with debris. 
The main findings are as follows:

\begin{itemize}
    \item Separating roof delineation from damage condition recognition preserves the roof where damage is most severe. A destroyed roof resembles the surrounding debris, and a single model trained for both tasks found no roof at all in several fully damaged tiles. On average, its damaged area and damage ratio deviated about 1.7 times as much as those of the developed two-phase framework. A model that only delineates roofs learns from every roof, whatever its condition.
    \item Mask2Former, which labels whole regions, measured damage best among the models tested for condition recognition. With the same roof outlines, the average deviation of the damage ratio was about 1.5 times as large with CLIPSeg and about twice as large with U-Net++. CLIPSeg, with frozen general-purpose features, mislabeled shadowed roof as damaged, and the pixel-by-pixel labels of U-Net++ tend to break roofs into fragments.
    \item Attention maps demonstrate how Mask2Former separates adjoining roof conditions: the query behind each condition, a learned slot that proposes one region, sharpens from coarse to fine. On one roof, the query for the tarp settled at once, and those for shadow and intact roofing tightened step by step. The roof delineation fixes where the roof is, and the
    queries decide what each part of it is.
    \item The framework successfully transfers to a different hazard and building style with new examples, without changing the architectural design. On Hurricane Melissa imagery from Jamaica, the Joplin-trained models overextended roofs onto adjoining surfaces such as paved terraces in every test tile. Fine-tuning with 95 and 35 labeled tiles for the two phases reduced the average error of the damaged roof area by 87\%.
\end{itemize}

To the authors' knowledge, this is the first framework to segment entire roofs at the pixel level into intact, damaged, tarp-covered, and background areas from post-event imagery alone, which turns roof damage from a label on a building into an area and a ratio measured on the roof. More broadly, the work sheds light on how imagery already collected within days of a wind event can deliver regional-scale, quantitative damage information for response and recovery, and on why locating a structure before judging its condition matters when damage erases its outline.

\section*{Data availability}
Data will be made available on request and after the manuscript is accepted.

\section*{Acknowledgements}
The authors appreciate the support from the National Science Foundation under award numbers EEC 2435462 and the Wind Hazard and Infrastructure Performance Center (WHIP-C) award WHIP2025-02. Any opinions, findings, conclusions or recommendations expressed in this study are those of the authors and do not necessarily reflect the views of the National Science Foundation. The authors also acknowledge the Startup Funds from Texas Tech University (TTU) and the high performance computing resources from the Texas Advanced Computing Center (TACC) at The University of Texas at Austin.

\section*{CRediT authorship contribution statement}
\noindent
\textbf{Weihe Sun}: Methodology, Software, Validation, Formal analysis, Investigation, Writing – original draft, Writing – review \& editing, Visualization, Supervision.
\textbf{Chuan Yu}: Software, Validation, Formal analysis, Investigation, Data curation, Writing – original draft, Visualization.
\textbf{Kevin Zhao}: Software, Data curation.
\textbf{Eshan Shrestha}: Software, Data curation.
\textbf{Guangzhao Chen}: Data curation, Writing – review \& editing.
\textbf{Douglas A. Smith}: Writing – review \& editing, Funding acquisition.
\textbf{Yimin Lu}: Conceptualization, Methodology, Validation, Formal analysis, Investigation, Resources, Writing – review \& editing, Visualization, Supervision, Project administration, Funding acquisition.

\bibliography{references}

\end{document}